\documentclass{article}
\usepackage{iclr2026_conference,times}

\usepackage{amsmath,amsfonts,bm}

\def\eqref#1{equation~\ref{#1}}

\def\1{\bm{1}}

\DeclareMathAlphabet{\mathsfit}{\encodingdefault}{\sfdefault}{m}{sl}
\SetMathAlphabet{\mathsfit}{bold}{\encodingdefault}{\sfdefault}{bx}{n}

\usepackage{url}
\usepackage{booktabs}
\usepackage{graphicx}
\usepackage{amsmath,amssymb}
\usepackage{multirow}
\usepackage{array}
\usepackage{xcolor}
\usepackage{float}
\usepackage{hyperref}
\hypersetup{hidelinks}

\newcolumntype{P}[1]{>{\raggedright\arraybackslash}p{#1}}
\newcommand{\pp}{\,\mathrm{pp}}

\iclrfinalcopy

\title{Answer First, Reason Later:\\[2pt]
When Commitment Order Costs Accuracy\\[2pt]
in Diffusion Language Models}

\author{%
Jewon Yeom$^{1}$ \quad
Jaewon Sok$^{2}$ \quad
Seonghyeon Park$^{3}$ \\[2pt]
\bfseries Jeongjae Park$^{1}$ \quad
Hwiyeong Lee$^{1}$ \quad
Taesup Kim$^{1,}$\thanks{\ \ Corresponding author.} \\[4pt]
{\normalfont\small $^{1}$Graduate School of Data Science, Seoul National University}\\
{\normalfont\small $^{2}$Department of Rural Systems Engineering, Seoul National University}\\
{\normalfont\small $^{3}$Department of Aerospace Engineering, Seoul National University}
}

\begin{document}
\maketitle
\lhead{}

\begin{abstract}
Masked diffusion language models revise many masked output positions in parallel. We call a token \emph{committed} once it becomes visible and is never masked again, and call a response \emph{answer-first} when the final answer commits before the reasoning printed ahead of it. On 1,069 GSM8K test questions, an explicit step-by-step instruction increases the accuracy difference between unrestricted decoding and a decoder that permits commitment only near the left-most unresolved position; unrestricted decoding also produces more answer-first trajectories. On MATH-500, the two LLaDA models spend most of a short output canvas on reasoning that commits after the answer, and the benefit of frontier gating decreases as that post-answer writing disappears. Dream-7B has little post-answer writing and follows a different accuracy pattern. A controlled four-option task reserves a one-token answer position before generation. Delaying that position outperforms an equally timed reasoning-token delay on LLaDA-8B, LLaDA-1.5, and Dream-7B. The raw difference is largest on Dream, whose free accuracy on the controlled task is lower. Answers commit much earlier under the reserved-position interface than in ordinary free-form generation, which limits how far the intervention result can be generalized. Commitment order affects the context used to complete a response and the allocation of a finite output canvas.
\end{abstract}

\section{Introduction}
\label{sec:intro}

Autoregressive language models expose one token at a time, so displayed order also fixes the order in which generated tokens can condition later text. Masked diffusion language models remove that constraint. They begin with a set of masked output positions---an \emph{output canvas}---and repeatedly replace selected masks with predicted tokens \citep{austin2021d3pm, nie2025llada}. Parallel and blockwise decoding exploit this flexibility for speed and infilling \citep{arriola2025block, arriola2026set, zuo2026window}. The same flexibility creates a reasoning problem that does not occur in left-to-right generation: a final answer printed at the end of a response may become permanent before the rationale printed before it.

This hidden order can change the response in two ways. Once an answer token becomes permanent, it is visible to later denoising steps and may condition text that is still masked. The order also determines how a finite canvas is used. If the answer is fixed early, much of the remaining canvas may be spent completing text around a conclusion that can no longer change. These possibilities matter for chain-of-thought evaluation because the displayed rationale need not reveal what was already fixed when that rationale was generated \citep{wei2022chain, turpin2023unfaithful, lanham2023faithfulness, boppana2026theater}.

We measure this order from the decoding trajectory. A token is \emph{committed} at the first denoising step after which it remains visible and is never masked again. A \emph{commitment policy} determines which masked positions are allowed to become permanent at each step. We call a response \emph{answer-first} when the answer commits before the median token in the reasoning region displayed ahead of it. Reasoning positions committed after the answer are called \emph{backfill}.

\begin{figure}[t]
\centering
\includegraphics[width=0.86\textwidth]{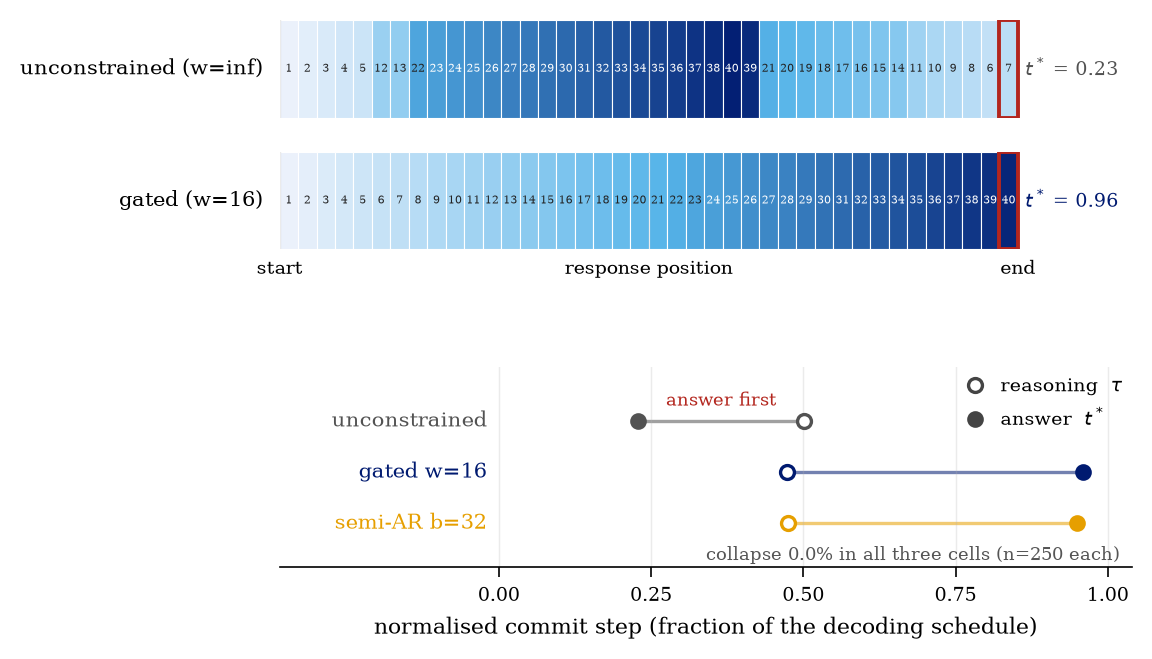}
\caption{\textbf{Displayed order and commitment order can differ.} Unrestricted decoding commits the answer seventh in the example, while frontier gating commits it last. The lower panel reports benchmark medians.}
\label{fig:teaser}
\end{figure}

Figure~\ref{fig:teaser} shows the event on MATH-500 \citep{hendrycks2021math, lightman2024verify}. It motivates three questions. Does an explicit reasoning instruction make accuracy more sensitive to the commitment policy? When does restricting distant commitments help, and when does it become unnecessary? Does delaying the answer position itself change accuracy, after controlling for the timing and duration of the delay?

GSM8K \citep{cobbe2021gsm8k} addresses prompt sensitivity by comparing unrestricted, frontier-gated, and semi-autoregressive decoding. MATH-500 addresses budget dependence through several canvas lengths and three models with different training histories. A controlled four-option task addresses the answer position directly by reserving it before generation and delaying only its availability. The first two analyses describe ordinary free-form generation. The controlled task changes the output interface, so it is used to identify a local causal effect rather than to estimate open-ended GSM8K accuracy.

The comparisons show that model and interface both matter. Explicit reasoning increases answer-first commitment under unrestricted GSM8K decoding. The two LLaDA models backfill most of a short MATH canvas, whereas Dream-7B usually commits its MATH answer after the reasoning. Once a one-token answer slot is declared, however, all three models commit that slot very early, and delaying it improves accuracy relative to a matched reasoning-token delay. The contrast shows that premature commitment depends on how the answer is represented as well as on the model and task.

\section{Related work}
\label{sec:related}

\paragraph{Order restrictions in diffusion decoding.}
Masked diffusion models permit many token-generation orders \citep{austin2021d3pm, nie2025llada}. Block diffusion, set diffusion, and windowed decoding restrict this freedom to improve speed or generation quality \citep{arriola2025block, arriola2026set, zuo2026window}. Deferred Commitment Decoding uses a confidence-aware moving window \citep{shu2026deferred}, while \citet{ni2026flexibility} study how arbitrary-order generation can skip difficult logical branches. Our frontier rule is intentionally simple: the model scores the full canvas, but only positions near the left-most unresolved token may be selected. This lets us change positional availability without introducing a learned threshold or tuning a new decoder for benchmark performance.

\paragraph{Reasoning order and faithfulness.}
Answer pre-commitment in autoregressive models has been studied through biased contexts, chain-of-thought perturbations, probes, and answer-conditioned rationales \citep{turpin2023unfaithful, lanham2023faithfulness, boppana2026theater, matcha2025, parekh2026drop}. A diffusion trajectory supplies a literal observation: once an answer token is visible and never masked again, later text is generated with that token in context. \citet{yu2026thinking} manipulate the requested display order and show that difficult answers can stabilize after reasoning even when printed first. We keep the display format fixed and alter the positions that may commit. The two studies probe different determinants of order: token certainty and positional availability.

\paragraph{Revision and termination.}
Remasking and overwrite methods allow a token to change after an earlier prediction \citep{wang2025remdm, yu2026revise}. Our controlled intervention acts before the answer is selected by withholding its position for a fixed period. Work on EOS overflow and variable-length generation concerns a separate termination failure \citep{kim2025rainbow, liu2026voidpadding, yang2025dllmvar, yang2026rhoeos}. The main experiments keep the released EOS/EOT handling fixed; Appendix~\ref{app:suppression} changes it only to show how termination settings can alter commitment order.

\section{Methods}
\label{sec:methods}
\label{sec:protocol}

\subsection{Models and tasks}

The experiments use three instruction-tuned diffusion language models. LLaDA-8B-Instruct is pretrained from scratch with a masked-diffusion objective and then supervised fine-tuned \citep{nie2025llada}. LLaDA-1.5 starts from the LLaDA family and adds VRPO preference optimization \citep{zhu2025llada15}. Dream-7B is initialized from an autoregressive language model and adapted to discrete diffusion with context-dependent noise scheduling \citep{ye2025dream}. These differences are useful for comparison, although they do not isolate any one training choice.

We evaluate grade-school arithmetic on GSM8K \citep{cobbe2021gsm8k}, competition mathematics on MATH-500 \citep{hendrycks2021math, lightman2024verify}, and Python generation on HumanEval \citep{chen2021humaneval} and MBPP \citep{austin2021mbpp}. Unless noted otherwise, the output canvas has $L{=}512$ positions, decoding uses 512 denoising steps at temperature zero, and the released low-confidence remasking procedure is used with EOS/EOT confidence suppression enabled.

The first 250 GSM8K test questions were used to establish the prompt comparison and to develop the trajectory analyses. Three questions inspected during interface development are excluded, leaving $n{=}247$. The same comparison and analysis code were then applied to the remaining 1,069 test questions, which supply the main estimate. We report the two subsets separately because the analysis was refined after the first subset was examined. The MATH experiments use both 250-question halves of MATH-500 and canvas lengths from 256 to 1024.

\subsection{Commitment policies}

Unrestricted decoding allows every masked position to commit. Frontier gating allows only positions $i<f(t)+w$, where $f(t)$ is the left-most mask at denoising step $t$; the default width is $w{=}16$. Semi-autoregressive decoding completes blocks of 32 positions from left to right. The permitted positions form the \emph{eligible set}. The model, prompt, temperature, denoising budget, number of commitments per step, and token scores are unchanged. A committed token stays visible during subsequent denoising steps. The log records this visible irreversible order; it does not reveal the order of latent computations inside the network.

\subsection{Trajectory measurements}

For position $i$, let $\tau(i)$ be its normalized first irreversible commitment step. We write $t^{*}$ for the median commitment step of the extracted answer span and $\tau$ for the median step of the preceding reasoning region. The order gap is
\begin{equation}
    g=t^{*}-\tau,
\end{equation}
so $g<0$ denotes answer-first commitment. Answer spans are located with a fixed extraction rule and mapped to tokenizer positions. Answer-only collapse denotes an extracted answer with essentially no non-EOS content before it. Correction analyses define answer-first status from the unrestricted trajectory before comparing policies.

For MATH, let $R$ be the reasoning positions. We measure
\begin{equation}
 p_{\mathrm{pre}}=\frac{|\{i\in R:\tau(i)<t^{*}\}|}{L},
 \qquad
 b_{\mathrm{backfill}}=\frac{|\{i\in R:\tau(i)>t^{*}\}|}{|R|}.
\end{equation}
The first quantity is the share of the canvas committed as reasoning before the answer. Backfill is the share of reasoning positions committed afterward. Neither quantity assesses whether a reasoning token is logically valid.

\subsection{Prompt--policy comparison}

GSM8K uses three instructions: direct answer, a format-only instruction that permits reasoning, and an explicit instruction that adds ``Let's think step by step.'' The central prompt--policy interaction is
\begin{equation}
I_{\mathrm{prompt}}=
(A_{\mathrm{explicit,gate}}-A_{\mathrm{explicit,uncon}})
-
(A_{\mathrm{format,gate}}-A_{\mathrm{format,uncon}}).
\label{eq:iprompt}
\end{equation}
The initial comparison defined this quantity. The answer-first shift and correction analysis were then applied without modification to the remaining 1,069 questions.

\subsection{Controlled answer delay}
\label{sec:intervention-method}

The answer-specific experiment uses a GSM8K-derived four-option task. For each question, three distinct distractors are drawn from GSM8K training answers with the same order of magnitude as the correct value, excluding the correct value itself. A fixed per-question permutation assigns the four values to A, B, C, and D. Each label is one tokenizer token. The distractors prevent scale alone from identifying the answer, but they are not intended to model common arithmetic mistakes. We use the same fixed sample of 640 questions for each model; its accuracy is not comparable with open-ended GSM8K accuracy.

Before generation, the canvas is initialized with an immutable ``Reasoning:'' delimiter, 280 masked reasoning positions, an immutable ``Final answer:'' delimiter, one masked answer position, and a fixed termination tail. There are 281 generated positions. The selected position is withheld until step 140, approximately halfway through this schedule. The free evaluation records the natural commitment time of the answer and every reasoning token. For the comparison condition, we choose a reasoning token that would have committed before step 140 and whose free commitment time is closest to the answer's. Selection uses commitment time only; correctness, the gold label, and intervention outcomes are not consulted.

Answer delay removes the answer position from the eligible set until step 140. The matched reasoning delay removes the selected reasoning position for the same interval. Both conditions change one position for the same duration. The intervention assigns negative-infinite commitment confidence to that position; it supplies no answer content and leaves the number of commitments per step unchanged. A valid timing match is available for all 640 questions on all three models. The median and 90th-percentile timing mismatch are one denoising step.

\subsection{Scoring and uncertainty}

GSM8K uses normalized numeric exact match. MATH combines normalized string equality with symbolic verification, applied identically across models and policies. The controlled task scores the final option label. Primary intervals are paired item-bootstrap intervals from 10,000 resamples. For selected paired comparisons we also report exact McNemar tests. This test uses only questions on which the two conditions disagree; Appendix~\ref{app:statistics} gives the calculation.

\section{Experimental results}
\label{sec:results}
\label{sec:interaction}

\subsection{Does explicit reasoning make commitment policy matter?}

\begin{figure}[H]
\centering
\includegraphics[width=0.98\textwidth]{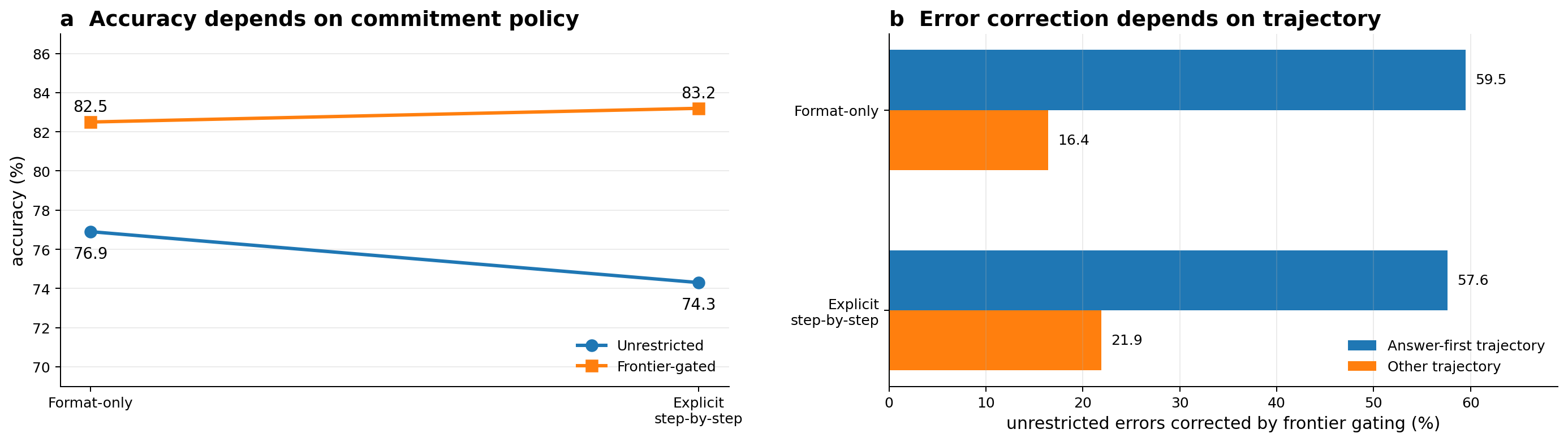}
\caption{\textbf{Prompt--policy interaction on the remaining 1,069 GSM8K test questions.} Left: accuracy under the format-only and explicit instructions. Right: among errors made by unrestricted decoding, the share corrected by frontier gating, grouped by the unrestricted trajectory's answer-first status.}
\label{fig:prompt-interaction}
\end{figure}

On the first 247 GSM8K questions, the prompt--policy interaction is $+7.7\pp$ with a 90\% CI of $[+3.6,+11.7]$ (Appendix~\ref{app:factorial}). The estimate from the remaining 1,069 questions is smaller but retains the same sign and decision. Figure~\ref{fig:prompt-interaction} reports this larger evaluation.

Format-only accuracy is $0.769$ under unrestricted decoding and $0.825$ under frontier gating. With the explicit instruction, the corresponding accuracies are $0.743$ and $0.832$. Equation~\ref{eq:iprompt} is therefore $+3.3\pp$ with a 90\% CI of $[+1.5,+5.1]$. The initial estimate was larger, so we use the 1,069-question result as the main effect size. The explicit instruction changes unrestricted accuracy by $-2.6\pp$ $[-4.4,-0.8]$ and gated accuracy by $+0.7\pp$ $[-0.9,+2.3]$. The supported result is the interaction: the instruction makes accuracy more dependent on where commitments are allowed.

The trajectory changes in the same direction. Explicit elicitation raises the unrestricted answer-first rate from $15.9\%$ to $24.5\%$ ($+8.6\pp$, $p{=}5.5\times10^{-17}$). The prompt difference in the continuous order gap is $0.0607$ with a 90\% CI of $[0.050,0.072]$.

To ask where frontier gating helps, we restrict the analysis to questions that unrestricted decoding gets wrong. An \emph{answer-first error} has $g<0$ in that unrestricted trajectory. Frontier gating corrects $59.5\%$ of format-only answer-first errors and $16.4\%$ of the other unrestricted errors (risk ratio $3.63$; exact odds ratio $7.41$ $[3.91,14.53]$). The paired difference in policy benefit between the two groups is $+35.4\pp$ $[+26.6,+44.2]$. The explicit instruction gives the same pattern (Appendix~\ref{app:factorial}). Because answer-first status is observed rather than assigned, this result locates the benefit but does not establish mediation.

The direct-answer instruction provides a boundary case. All three policies generate identical tokens, commitment steps, and per-step commitment counts on the first 250 questions, and a fresh 20-question evaluation reproduces the result. Their eligible sets are nevertheless different: unrestricted decoding exposes 256.5 positions per step on average, whereas frontier gating exposes 15.8. The restriction is active but does not bind because the response contains seven non-special tokenizer tokens and remains inside the local window. Without a separately generated reasoning region, there is no long-range order for the policies to change.

\subsection{When does restricting commitment help on MATH-500?}
\label{sec:canvas}

At $L{=}512$, unrestricted LLaDA-8B commits the answer at $t^{*}{=}0.229$ and the median supporting token at $\tau{=}0.501$; $73.9\%$ of responses are answer-first. Frontier gating improves rescored accuracy by $+8.1\pp$ $[+3.6,+12.6]$. The extra step-by-step sentence does not produce a resolved prompt--policy interaction on any of the three models: $-1.2\pp$ $[-6.1,+3.6]$ for LLaDA-8B, $-0.4\pp$ $[-5.3,+4.5]$ for LLaDA-1.5, and $0.0\pp$ $[-4.9,+4.5]$ for Dream-7B. The MATH result therefore concerns the decoding policy and canvas budget, not sensitivity to this prompt sentence.

\begin{figure}[H]
\centering
\includegraphics[width=0.99\textwidth]{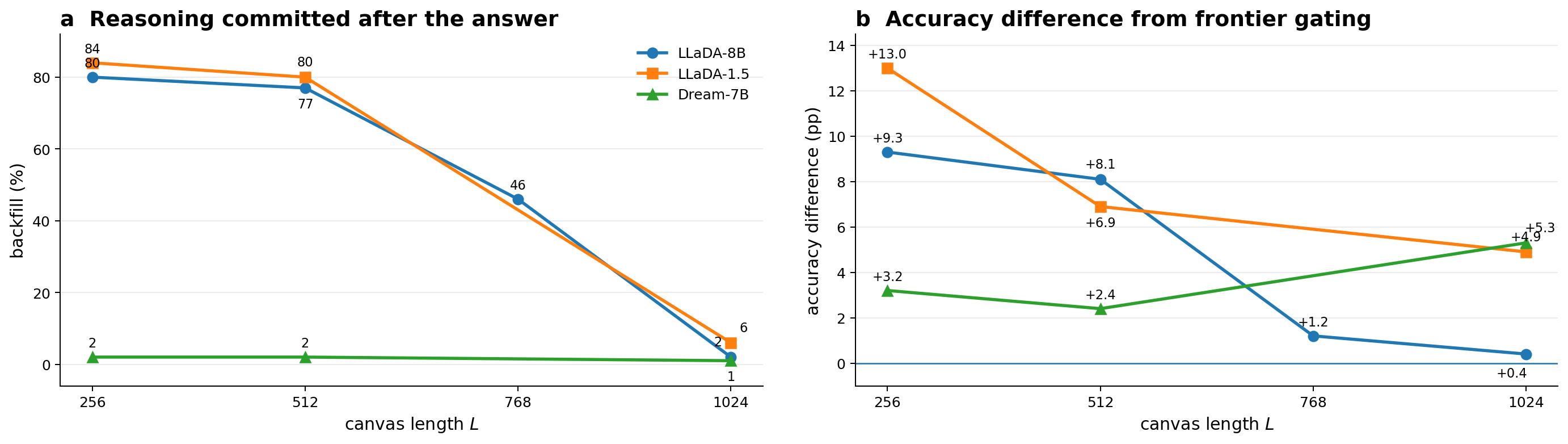}
\caption{\textbf{MATH results across canvas lengths.} Backfill is the fraction of reasoning positions committed after the answer. The right panel shows frontier-gated minus unrestricted accuracy. The $L{=}768$ point was collected only for LLaDA-8B.}
\label{fig:canvas-allocation}
\end{figure}

Figure~\ref{fig:canvas-allocation} compares post-answer writing and accuracy across canvas lengths. Only LLaDA-8B was evaluated at $L{=}768$; LLaDA-1.5 and Dream-7B were evaluated at $L\in\{256,512,1024\}$.

The two LLaDA models show similar budget dependence. At $L{=}256$ and $512$, unrestricted decoding commits only $15$--$22\%$ of the canvas as reasoning before the answer; $77$--$84\%$ of the reasoning positions commit afterward. Frontier gating commits $95$--$96\%$ of these short canvases before the answer. As the canvas grows, unrestricted backfill falls to $2\%$ for LLaDA-8B and $6\%$ for LLaDA-1.5. Their accuracy gains decrease from $+9.3$ to $+0.4\pp$ and from $+13.0$ to $+4.9\pp$, respectively.

Dream-7B behaves differently. Its unrestricted backfill is already $2\%$ at $L{=}256$ and $512$ and $1\%$ at $L{=}1024$, so there is little post-answer canvas use for frontier gating to remove. The accuracy differences are $+3.2$, $+2.4$, and $+5.3\pp$. At $L{=}1024$, 54 unrestricted responses fill the canvas and all are incorrect. Excluding those responses yields smaller differences whose intervals include zero, but the accuracy pattern still does not become monotonic. Backfill alone therefore does not explain the Dream results.

The LLaDA-8B distribution also changes shape with the canvas. The fraction of unrestricted trajectories with less than $10\%$ backfill rises from $4.1\%$ at $L{=}256$ to $20.7\%$, $42.9\%$, and $57.3\%$ at $L{=}512,768,1024$. Short canvases contain many trajectories that commit the answer early and continue writing until the canvas is full; longer canvases contain a larger group that terminates with little backfill. A single median obscures this change in composition.

At the measured points, frontier gating with half the canvas matches or exceeds unrestricted decoding with twice the canvas. LLaDA-8B scores $0.296$ with gated $L{=}256$ versus $0.275$ with unrestricted $L{=}512$, and $0.356$ with gated $L{=}512$ versus $0.348$ with unrestricted $L{=}1024$. These are point comparisons, not equivalence tests. Full model-by-length results appear in Appendix~\ref{app:math-budget}.

\subsection{Does delaying the answer itself change accuracy?}
\label{sec:causal}

The GSM8K analysis identifies the trajectories on which frontier gating helps, but answer-first status is observational. The controlled experiment changes the availability of the declared answer position. The task presents four numerical options, reserves a one-token A/B/C/D slot, and compares two delays of equal width and duration.

A free evaluation records when the answer and each reasoning token would naturally commit. Answer delay blocks the answer slot until step 140. The matched reasoning delay blocks a reasoning token whose free commitment time is nearest to the answer's and that would otherwise commit before release. This comparison tests whether an equally early delay helps at any position, and whether the answer position has a distinct effect.

\begin{figure}[H]
\centering
\includegraphics[width=0.99\textwidth]{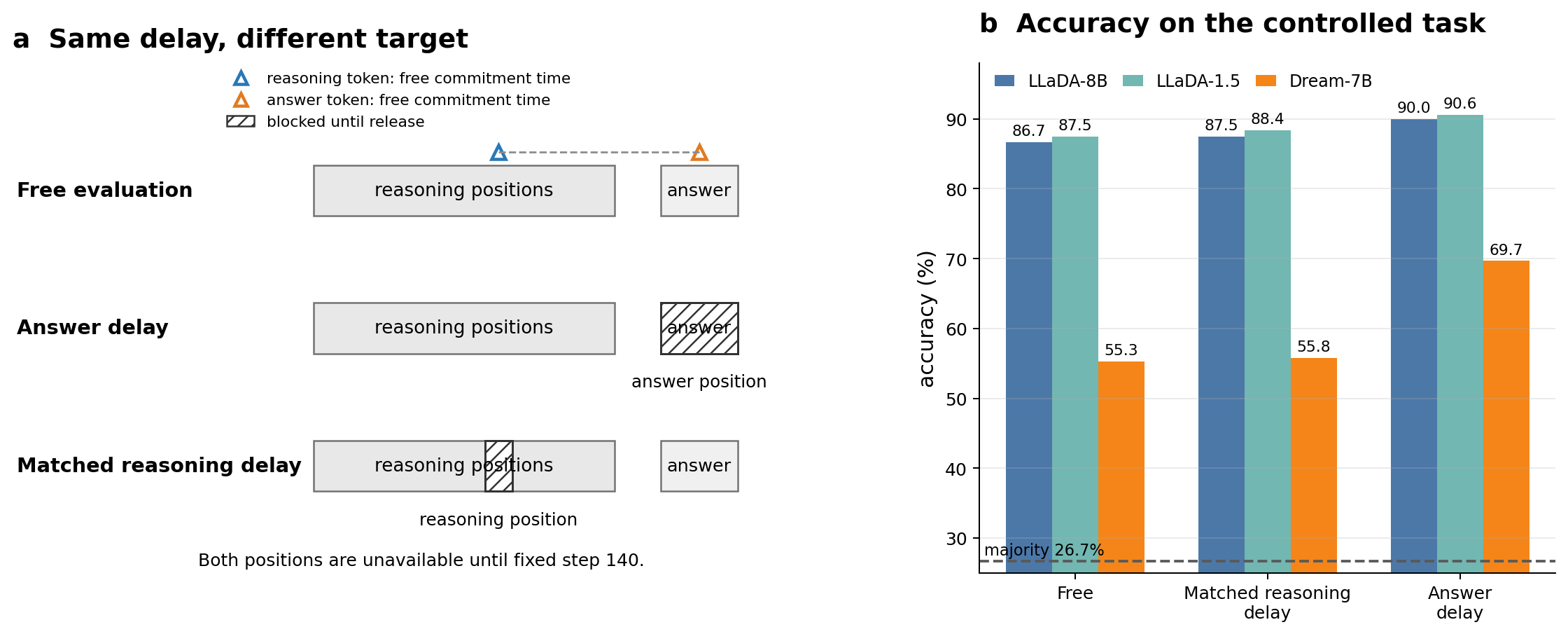}
\caption{\textbf{Controlled answer delay.} Panel a shows the two delays, which share the same target width and release step. Panel b reports accuracy. Answer delay exceeds the matched reasoning delay by $+2.5\pp$ $[+1.1,+4.1]$ on LLaDA-8B, $+2.2\pp$ $[+0.6,+3.9]$ on LLaDA-1.5, and $+13.9\pp$ $[+11.1,+16.7]$ on Dream-7B.}
\label{fig:answer-delay}
\end{figure}

For LLaDA-8B, free, matched-reasoning, and answer-delay accuracies are $0.867$, $0.875$, and $0.900$. The answer-minus-reasoning contrast is $+2.5\pp$ $[+1.1,+4.1]$; the 95\% CI is $[+0.8,+4.4]$ and exact McNemar $p{=}0.009$. LLaDA-1.5 gives $0.875$, $0.884$, and $0.906$, with a contrast of $+2.2\pp$ $[+0.6,+3.9]$ and McNemar $p{=}0.044$. Dream-7B gives $0.553$, $0.558$, and $0.697$, with a contrast of $+13.9\pp$ $[+11.1,+16.7]$ and McNemar $p{=}5.25\times10^{-15}$.

The Dream difference is much larger in raw percentage points, but Dream also begins with substantially lower free accuracy on this controlled task. Relative to the remaining distance from free accuracy to 1.0, answer delay recovers $24.8\%$, $24.8\%$, and $32.2\%$ for LLaDA-8B, LLaDA-1.5, and Dream-7B. These three ratios are descriptive. They prevent a raw cross-model difference from being read as a direct measure of how premature each model is.

The reasoning delay is not a weaker perturbation. It displaces a median of 115 top-$k$ selections versus 81.5 for answer delay on LLaDA-8B, 123 versus 105 on LLaDA-1.5, and 121 versus 77 on Dream-7B. Answer delay changes the selected option more often and the changes favor correction: correction/harm rates are $4.5/1.2\%$, $4.8/1.7\%$, and $18.0/3.6\%$ across the three models. Full intervention diagnostics appear in Appendix~\ref{app:a2}.

The scaffold changes the generated trajectory. On the controlled task, normalized answer commitment is $0.02$ for LLaDA-8B and $0.007$ for Dream-7B, compared with $0.23$ and $0.42$ in their free-form MATH evaluations. Dream therefore moves from little backfill and relatively late answers on MATH to an answer fixed within the first few scaffolded steps. The task and interface change together, so this comparison is diagnostic rather than a matched estimate of a scaffold effect. All three intervention conditions use the same scaffold, preserving the internal answer-versus-reasoning contrast while limiting its generalization to ordinary free-form generation.

\subsection{Code generation as a structural control}
\label{sec:code}

A numerical response ends with a compact answer span that can become permanent separately from its explanation. A generated Python function has no equally clean separation: the body and the returned value jointly constitute the answer. Under the same EOS/EOT handling used in the main experiments, frontier gating changes HumanEval accuracy by $-1.2\pp$ and MBPP by $-0.4\pp$; neither difference is resolved from zero. These results do not show that commitment order is irrelevant to code. They show that the present frontier restriction has no measured benefit on two tasks where the output is less separable. A syntax-aware timing analysis would need to track function signatures, control flow, and return expressions under matched generation settings; the current return-token diagnostic is kept in Appendix~\ref{app:tasks} because its termination setting differs.

\section{Discussion}
\label{sec:discussion}

The GSM8K results concern visible conditioning order. An explicit reasoning instruction increases the frequency with which unrestricted decoding fixes the answer before the displayed rationale. Frontier gating helps most on errors with that trajectory. The controlled task then changes the answer position directly: withholding it produces a larger accuracy change than withholding an equally early reasoning position on all three models. These measurements concern irreversible visible tokens. They do not reveal when the network completed an internal calculation.

Early commitment should not be interpreted as guessing from a marginal prior. Every denoising step conditions on the full question and current canvas. An answer committed after a few steps may reflect several full forward passes even though readable reasoning has not stabilized. The result concerns the point at which the answer becomes unavailable for revision.

The MATH comparison adds a budget effect. Both LLaDA models devote most of a short canvas to text that commits after the answer, and the benefit of frontier gating decreases as that post-answer writing disappears. LLaDA-1.5 adds VRPO preference optimization to the LLaDA family, yet retains a similar allocation pattern. Dream-7B, initialized from an autoregressive model before diffusion adaptation, has little backfill in free-form MATH. This contrast is consistent with training history affecting commitment order, but the three models also differ in architecture, data, and post-training.

The controlled task also shows that model lineage alone does not determine the trajectory. Dream commits answers late in free-form MATH, then commits almost immediately when a fixed answer slot is provided. LLaDA-1.5 and Dream also have similarly early median answer times under the scaffold, while their raw answer-delay gains differ substantially. Output format and available headroom therefore change the observed trajectory and effect size. The three-model intervention establishes a causal effect for an explicit answer interface; it does not rank the models by a context-independent tendency toward premature commitment.

The code results provide a further boundary. A numerical conclusion occupies a small span that can become permanent separately from its derivation. A function body does not divide as cleanly into support and answer, and the frontier restriction produces no measured gain on HumanEval or MBPP. Commitment order may still matter for signatures, control-flow skeletons, or return expressions; those units require syntax-aware measurements under matched generation settings.

A general rule favoring later answers is inconsistent with the data. Once global reachability is removed, changing the frontier width does not produce a monotonic accuracy response. Short direct answers are unchanged because the restriction never excludes a selected position. Dream-7B and the larger MATH canvases show that the same positional rule can have little benefit when post-answer canvas use is already low. Selective delay is most relevant when a compact conclusion is independently reachable and early commitment removes revision opportunity or consumes scarce output space.

\section{Limitations}
\label{sec:limitations}

The GSM8K prompt--policy interaction is established on LLaDA-8B. The MATH canvas comparison and controlled answer-delay experiment include LLaDA-8B, LLaDA-1.5, and Dream-7B. The two LLaDA answer-delay effects are smaller than the $5\pp$ value used for power planning. Dream shows a larger raw difference, but its free accuracy on the controlled task is much lower, so absolute effect sizes should not be compared as model-level measures of prematurity.

The four-option intervention measures a controlled mechanism; its accuracy is not open-ended GSM8K accuracy. Its distractors match the correct answer's order of magnitude but come from other training questions and do not represent problem-specific arithmetic errors. The preallocated answer slot also changes commitment behavior. LLaDA-8B and Dream-7B commit much earlier in the scaffolded task than in free-form MATH. Since task and interface change together, this comparison does not isolate a pure scaffold effect. The answer-delay contrast remains internally matched because all conditions share the same scaffold, but generalization to unconstrained generation is limited.

The answer position may matter because it closes the problem, or because it is an unusually influential token. The timing-matched control does not match semantic influence. A stronger specificity test would delay other high-impact positions. Dream's free accuracy of $0.553$ on the controlled task also remains unexplained: it has no option-format failures and performs best among the three models on the reported MATH condition, so neither format compliance nor early timing alone accounts for the gap.

The MATH model-by-length conditions reuse the same questions, making cross-length associations descriptive. The $L{=}768$ condition was collected only for LLaDA-8B. The $L{=}1024$ Dream condition contains 54 unrestricted responses that fill the canvas and are all incorrect; this is reported separately rather than folded into a general backfill account. HumanEval and MBPP give a broad structural comparison only. More informative code experiments would track syntax-aware regions under the same termination setting. The main experiments retain the released EOS/EOT suppression; removing it exposes a separate termination failure (Appendix~\ref{app:suppression}).

\section{Conclusion}
\label{sec:conclusion}

A diffusion language model can make a final answer permanent while the reasoning printed before it remains unresolved. On GSM8K, explicit reasoning instructions make accuracy more sensitive to the commitment policy, and frontier gating helps most on answer-first errors. On MATH, the benefit depends on finite-canvas use and differs across model families. In a controlled interface with a declared answer slot, delaying that slot improves accuracy over a matched reasoning-position delay on all three models. The scaffold itself induces earlier answer commitment, which narrows the causal claim to that interface. These results support selective deferral of compact conclusions when early commitment removes revision time or consumes scarce output space.

\section*{AI use statement}
Generative AI tools assisted with experimental-design feedback, implementation, statistical checks, figures, and language editing. The authors reviewed the resulting code and text, recomputed the reported statistics from saved outputs, and verified the cited sources.

\bibliography{references}

@inproceedings{austin2021d3pm,
  title={Structured denoising diffusion models in discrete state-spaces},
  author={Austin, Jacob and Johnson, Daniel D and Ho, Jonathan and Tarlow, Daniel and van den Berg, Rianne},
  booktitle={Advances in Neural Information Processing Systems},
  year={2021}
}

@article{nie2025llada,
  title={Large Language Diffusion Models},
  author={Nie, Shen and Zhu, Fengqi and You, Zebin and Zhang, Xiaolu and Ou, Jingyang and Hu, Jun and Zhou, Jun and Lin, Yankai and Wen, Ji-Rong and Li, Chongxuan},
  journal={arXiv preprint arXiv:2502.09992},
  year={2025}
}

@inproceedings{turpin2023unfaithful,
  title={Language Models Don't Always Say What They Think: Unfaithful Explanations in Chain-of-Thought Prompting},
  author={Turpin, Miles and Michael, Julian and Perez, Ethan and Bowman, Samuel R},
  booktitle={Advances in Neural Information Processing Systems},
  year={2023}
}

@article{lanham2023faithfulness,
  title={Measuring Faithfulness in Chain-of-Thought Reasoning},
  author={Lanham, Tamera and others},
  journal={arXiv preprint arXiv:2307.13702},
  year={2023}
}

@article{cobbe2021gsm8k,
  title={Training Verifiers to Solve Math Word Problems},
  author={Cobbe, Karl and others},
  journal={arXiv preprint arXiv:2110.14168},
  year={2021}
}

@inproceedings{hendrycks2021math,
  title={Measuring Mathematical Problem Solving With the MATH Dataset},
  author={Hendrycks, Dan and others},
  booktitle={NeurIPS Datasets and Benchmarks},
  year={2021}
}

@inproceedings{lightman2024verify,
  title={Let's Verify Step by Step},
  author={Lightman, Hunter and others},
  booktitle={International Conference on Learning Representations},
  year={2024},
  note={MATH-500 subset origin}
}

@article{kim2025rainbow,
  title={Rainbow Padding: Mitigating Early Termination in Instruction-Tuned Diffusion {LLMs}},
  author={Kim, Bumjun and Jeon, Dongjae and Kim, Dueun and Jeung, Wonje and No, Albert},
  journal={arXiv preprint arXiv:2510.03680}, year={2025}}

@article{liu2026voidpadding,
  title={{VoidPadding}: Let {[VOID]} Handle Padding in Masked Diffusion Language Models so that {[EOS]} Can Focus on Semantic Termination},
  author={Liu, Chunyu and Fan, Zhengyang and Yang, Kaisen and Lamb, Alex},
  journal={arXiv preprint arXiv:2606.17999}, year={2026}}

@article{yang2026rhoeos,
  title={$\rho$-{EOS}: Training-free Bidirectional Variable-Length Control for Masked Diffusion {LLMs}},
  author={Yang, Jingyi and Jiang, Yuxian and Shao, Jing},
  journal={arXiv preprint arXiv:2601.22527}, year={2026}}

@article{yang2025dllmvar,
  title={Diffusion {LLM} with Native Variable Generation Lengths: Let {[EOS]} Lead the Way},
  author={Yang, Yicun and Wang, Cong and Wang, Shaobo and Wen, Zichen and Qi, Biqing and Xu, Hanlin and Zhang, Linfeng},
  journal={arXiv preprint arXiv:2510.24605}, year={2025}}

@article{zuo2026window,
  title={{Window-Diffusion}: Accelerating Diffusion Language Model Inference with Windowed Token Pruning and Caching},
  author={Zuo, Fengrui and Ke, Zhiwei and Liu, Yiming and Lou, Wenqi and Wang, Chao and Zhou, Xuehai},
  journal={arXiv preprint arXiv:2601.20332}, year={2026}}

@article{shu2026deferred,
  title={Deferred Commitment Decoding for Diffusion Language Models with Confidence-Aware Sliding Windows},
  author={Shu, Yingte and Tian, Yuchuan and Xu, Chao and Wang, Yunhe and Chen, Hanting},
  journal={arXiv preprint arXiv:2601.02076}, year={2026}}

@inproceedings{arriola2026set,
  title={Set Diffusion: Interpolating Token Orderings Between Autoregression and Diffusion for Fast and Flexible Decoding},
  author={Arriola, Marianne and Kuleshov, Volodymyr},
  booktitle={International Conference on Machine Learning},
  year={2026}}

@article{ni2026flexibility,
  title={The Flexibility Trap: Rethinking the Value of Arbitrary Order in Diffusion Language Models},
  author={Ni, Zanlin and Wang, Shenzhi and Yue, Yang and Yu, Tianyu and Zhao, Weilin and Hua, Yeguo and Chen, Tianyi and Song, Jun and Yu, Cheng and Zheng, Bo and Huang, Gao},
  journal={arXiv preprint arXiv:2601.15165}, year={2026},
  note={ICML 2026}}

@article{matcha2025,
  title={Robust Answers, Fragile Logic: Probing the Decoupling Hypothesis in {LLM} Reasoning},
  author={Jiang, Enyi and Xu, Changming and Singh, Nischay and Qiu, Tian and Singh, Gagandeep},
  journal={arXiv preprint arXiv:2505.17406}, year={2025}}

@article{parekh2026drop,
  title={Drop the Act: Probe-Filtered {RL} for Faithful Chain-of-Thought Reasoning},
  author={Parekh, Swapnil},
  journal={arXiv preprint arXiv:2605.11467}, year={2026}}

@article{ye2025dream,
  title={Dream {7B}: Diffusion Large Language Models},
  author={Ye, Jiacheng and Xie, Zhihui and Zheng, Lin and Gao, Jiahui and Wu, Zirui and Jiang, Xin and Li, Zhenguo and Kong, Lingpeng},
  journal={arXiv preprint arXiv:2508.15487}, year={2025}}

@article{boppana2026theater,
  title={Reasoning Theater: Disentangling Model Beliefs from Chain-of-Thought},
  author={Boppana, Siddharth and Ma, Andrew and Loeffler, Max and Sarfati, Raphael and Bigelow, Eric and Geiger, Atticus and Lewis, Owen and Merullo, Jack},
  journal={arXiv preprint arXiv:2603.05488}, year={2026}}

@inproceedings{arriola2025block,
  title={Block Diffusion: Interpolating Between Autoregressive and Diffusion Language Models},
  author={Arriola, Marianne and Gokaslan, Aaron and Chiu, Justin T and Yang, Zhihan and Qi, Zhixuan and Han, Jiaqi and Sahoo, Subham Sekhar and Kuleshov, Volodymyr},
  booktitle={International Conference on Learning Representations},
  year={2025}}

@article{yu2026revise,
  title={Revise, Don't Freeze: Sampler-Matched Training for Self-Correcting Masked Diffusion Language Models},
  author={Yu, Longxuan and Zhang, Shaorong and Fu, Yu and Liu, Hui and Dong, Yue and Ver Steeg, Greg},
  journal={arXiv preprint arXiv:2606.01026}, year={2026}}

@inproceedings{wang2025remdm,
  title={Remasking Discrete Diffusion Models with Inference-Time Scaling},
  author={Wang, Guanghan and Schiff, Yair and Sahoo, Subham Sekhar and Kuleshov, Volodymyr},
  booktitle={Advances in Neural Information Processing Systems},
  year={2025}}

@inproceedings{wei2022chain,
  title={Chain-of-Thought Prompting Elicits Reasoning in Large Language Models},
  author={Wei, Jason and Wang, Xuezhi and Schuurmans, Dale and Bosma, Maarten and Ichter, Brian and Xia, Fei and Chi, Ed H. and Le, Quoc V. and Zhou, Denny},
  booktitle={Advances in Neural Information Processing Systems},
  year={2022}}

@article{yu2026thinking,
  title={Thinking Out of Order: When Output Order Stops Reflecting Reasoning Order in Diffusion Language Models},
  author={Yu, Longxuan and Fu, Yu and Zhang, Shaorong and Liu, Hui and Varma T, Mukund and Ver Steeg, Greg and Dong, Yue},
  journal={arXiv preprint arXiv:2601.22035}, year={2026}}

@article{zhu2025llada15,
  title={{LLaDA 1.5}: Variance-Reduced Preference Optimization for Large Language Diffusion Models},
  author={Zhu, Fengqi and Wang, Rongzhen and Nie, Shen and Zhang, Xiaolu and Wu, Chunwei and Hu, Jun and Zhou, Jun and Chen, Jianfei and Lin, Yankai and Wen, Ji-Rong and Li, Chongxuan},
  journal={arXiv preprint arXiv:2505.19223},
  year={2025}
}

@article{chen2021humaneval,
  title={Evaluating Large Language Models Trained on Code},
  author={Chen, Mark and Tworek, Jerry and Jun, Heewoo and Yuan, Qiming and Pinto, Henrique Ponde de Oliveira and Kaplan, Jared and Edwards, Harri and Burda, Yuri and Joseph, Nicholas and Brockman, Greg and others},
  journal={arXiv preprint arXiv:2107.03374},
  year={2021}
}

@article{austin2021mbpp,
  title={Program Synthesis with Large Language Models},
  author={Austin, Jacob and Odena, Augustus and Nye, Maxwell and Bosma, Maarten and Michalewski, Henryk and Dohan, David and Jiang, Ellen and Cai, Carrie and Terry, Michael and Le, Quoc and Sutton, Charles},
  journal={arXiv preprint arXiv:2108.07732},
  year={2021}
}
\bibliographystyle{iclr2026_conference}

\clearpage
\appendix
\section{GSM8K prompt comparison}
\label{app:factorial}

\subsection{Instructions and manipulation checks}

The three instructions are reproduced below.
\begin{quote}\ttfamily\small
\textbf{Direct answer.} Respond with only the final answer in the form ``The answer is X''. Do not include any reasoning, explanation, or intermediate steps.\\[4pt]
\textbf{Format-only.} Finish your response with ``The answer is X'', where X is the final answer.\\[4pt]
\textbf{Explicit step-by-step.} Let's think step by step. Finish your response with ``The answer is X'', where X is the final answer.
\end{quote}

\begin{table}[H]
\centering
\caption{Manipulation checks on the first 250 GSM8K questions. Length is the median number of generated non-special tokens.}
\label{tab:manipulation}
\resizebox{\textwidth}{!}{%
\begin{tabular}{lccccc}
\toprule
Instruction & Reasoning present & Length & Declared format & Extraction & Degenerate \\
\midrule
Direct answer & 0.0\% & 4 & 100.0\% & 100.0\% & 0.0\% \\
Format-only & 96.4\% & 277 & 94.3\% & 100.0\% & 0.0\% \\
Explicit step-by-step & 100.0\% & 284 & 98.4\% & 100.0\% & 0.0\% \\
\bottomrule
\end{tabular}%
}
\end{table}

\subsection{First 247 questions}

\begin{table}[H]
\centering
\caption{Accuracy on the first GSM8K subset ($n{=}247$ paired after excluding three interface-development questions).}
\label{tab:factorial-grid}
\begin{tabular}{lccc}
\toprule
Instruction & Unrestricted & Gated ($w{=}16$) & Semi-AR (block 32) \\
\midrule
Direct answer & 0.457 & 0.457 & 0.457 \\
Format-only & 0.777 & 0.846 & 0.842 \\
Explicit step-by-step & 0.729 & 0.874 & 0.854 \\
\bottomrule
\end{tabular}
\end{table}

The prompt--policy interaction is $+7.7\pp$ with a 90\% paired-bootstrap CI of $[+3.6,+11.7]$. Replacing frontier gating with semi-autoregressive decoding gives $+6.1\pp$ $[+2.0,+10.1]$.

\begin{table}[H]
\centering
\caption{Explicit step-by-step minus format-only accuracy on the first GSM8K subset.}
\label{tab:simple-effects}
\begin{tabular}{lcccc}
\toprule
Policy & Difference & 90\% CI & Discordant pairs & Exact McNemar $p$ \\
\midrule
Unrestricted & $-4.9\pp$ & $[-8.9,-0.8]$ & 12/24 & 0.065 \\
Gated ($w{=}16$) & $+2.8\pp$ & $[-0.8,+6.5]$ & 17/10 & 0.248 \\
Semi-AR (block 32) & $+1.2\pp$ & $[-2.4,+4.9]$ & 16/13 & 0.711 \\
\bottomrule
\end{tabular}
\end{table}

\subsection{Remaining 1,069 questions}

\begin{table}[H]
\centering
\caption{Accuracy on GSM8K test questions 250--1318 ($n{=}1069$ paired).}
\label{tab:holdout-grid}
\begin{tabular}{lcc}
\toprule
Instruction & Unrestricted & Gated ($w{=}16$) \\
\midrule
Format-only & 0.769 & 0.825 \\
Explicit step-by-step & 0.743 & 0.832 \\
\bottomrule
\end{tabular}
\end{table}

The interaction is $+3.3\pp$ $[+1.5,+5.1]$. Among unrestricted errors, format-only correction rates are $59.5\%$ for answer-first items and $16.4\%$ for other items. Under explicit elicitation they are $57.6\%$ and $21.9\%$. The corresponding differences in policy benefit are $+35.4\pp$ $[+26.6,+44.2]$ and $+32.4\pp$ $[+25.2,+39.7]$.

\subsection{Direct-answer check}

All 250 direct-answer outputs are token-identical across the three policies, and their 512-position commitment-step arrays are identical. A fresh 20-question evaluation confirms equality of tokens, commitment steps, and per-step commit counts. Mean eligible-set sizes are 256.5 positions per step for unrestricted decoding, 15.8 for frontier gating, and 16.5 for semi-autoregressive decoding. The output contains seven non-special tokenizer tokens and stays inside each policy's reachable region.

\section{Statistical details}
\label{app:statistics}

Paired bootstrap intervals resample questions, preserving all conditions for a question within each resample. For McNemar's test, let $b$ be the number of questions correct only under condition A and $c$ the number correct only under condition B. Questions on which the conditions agree do not affect the test. Under the null hypothesis of equal paired accuracy, either condition is equally likely to be the sole correct one, so the exact two-sided $p$-value is computed from $b\sim\mathrm{Binomial}(b+c,1/2)$. Tables report discordant pairs as $b/c$.

\section{MATH order and canvas allocation}
\label{app:math-budget}

\subsection{Commitment order with a 512-position canvas}

\begin{table}[H]
\centering
\caption{LLaDA-8B MATH trajectories at $L{=}512$ ($n{=}249$ paired). Accuracy is reported in Table~\ref{tab:canvas-sweep}.}
\label{tab:mathorder}
\begin{tabular}{lcccc}
\toprule
Policy & Collapse & $t^{*}$ & $\tau$ & Answer-first \\
\midrule
Unrestricted & 0.0\% & 0.229 & 0.501 & 73.9\% \\
Gated ($w{=}16$) & 0.0\% & 0.960 & 0.473 & 0.0\% \\
Semi-AR (block 32) & 0.0\% & 0.949 & 0.476 & 0.0\% \\
\bottomrule
\end{tabular}
\end{table}

\begin{table}[H]
\centering
\caption{MATH prompt--policy interaction at $L{=}512$. Intervals that cross zero and extend beyond $\pm3\pp$ are inconclusive rather than evidence of equivalence.}
\label{tab:math-trigger-models}
\begin{tabular}{lcc}
\toprule
Model & $n$ & $I_{\mathrm{prompt}}$ (90\% CI) \\
\midrule
LLaDA-8B & 247 & $-1.2\pp$ $[-6.1,+3.6]$ \\
LLaDA-1.5 & 247 & $-0.4\pp$ $[-5.3,+4.5]$ \\
Dream-7B & 247 & $0.0\pp$ $[-4.9,+4.5]$ \\
\bottomrule
\end{tabular}
\end{table}

\subsection{Frontier width}
\label{app:window}

Across $w\in\{1,4,16,64,128\}$, answer timing changes while accuracy remains within overlapping uncertainty intervals. Every finite width remains above unrestricted decoding in the LLaDA-8B comparison. This result rules out a simple dose response in which progressively later answer commitment produces progressively higher accuracy.

\subsection{Models and canvas lengths}

\begin{table}[H]
\centering
\small
\setlength{\tabcolsep}{4pt}
\caption{MATH allocation and accuracy across models and canvas lengths. ``Pre'' is the fraction of the full canvas committed as reasoning before the answer. Gated backfill is 0\% in every listed condition. LLaDA-8B alone was evaluated at $L{=}768$.}
\label{tab:canvas-sweep}
\begin{tabular}{llrrrrl}
\toprule
Model & $L$ & Unres. backfill & Unres. pre & Gated pre & Gain & 90\% CI \\
\midrule
LLaDA-8B & 256  & 80\% & 0.19 & 0.96 & $+9.3\pp$ & $[+4.9,+13.8]$ \\
          & 512  & 77\% & 0.22 & 0.95 & $+8.1\pp$ & $[+3.6,+12.6]$ \\
          & 768  & 46\% & 0.29 & 0.66 & $+1.2\pp$ & $[-2.0,+4.5]$ \\
          & 1024 & 2\%  & 0.31 & 0.47 & $+0.4\pp$ & $[-3.2,+4.0]$ \\
\midrule
LLaDA-1.5 & 256  & 84\% & 0.15 & 0.96 & $+13.0\pp$ & $[+8.1,+17.8]$ \\
           & 512  & 80\% & 0.20 & 0.96 & $+6.9\pp$ & $[+2.0,+11.7]$ \\
           & 1024 & 6\%  & 0.25 & 0.47 & $+4.9\pp$ & $[+1.2,+8.5]$ \\
\midrule
Dream-7B & 256  & 2\% & 0.67 & 0.80 & $+3.2\pp$ & $[+0.4,+6.1]$ \\
          & 512  & 2\% & 0.40 & 0.51 & $+2.4\pp$ & $[-0.8,+5.7]$ \\
          & 1024 & 1\% & 0.21 & 0.23 & $+5.3\pp$ & $[+2.4,+8.5]$ \\
\bottomrule
\end{tabular}
\end{table}

At $L{=}1024$, Dream-7B unrestricted accuracy is reduced by 54 responses that fill the canvas and are all incorrect. Excluding this group gives gated-minus-unrestricted differences of $+2.1$, $+2.3$, and $+1.0\pp$ across $L{=}256,512,1024$; all three intervals include zero. This sensitivity analysis does not make the Dream pattern match the LLaDA pattern.

Across the ten measured model--canvas settings, unrestricted backfill and the gated-minus-unrestricted accuracy difference have a descriptive Pearson correlation of $r{=}0.76$ (90\% interval $[0.48,0.92]$). The settings reuse questions and models, and this summary was formed after inspecting the model-by-length results. It is reported as a descriptive association, not a hypothesis test.

For LLaDA-8B, the fraction of unrestricted trajectories with less than 10\% backfill is $4.1\%$, $20.7\%$, $42.9\%$, and $57.3\%$ for $L{=}256,512,768,1024$. These threshold counts show the shift toward near-zero backfill; a full histogram requires the saved per-question trajectories and is not reconstructed from the summary table.

\begin{table}[H]
\centering
\caption{Point comparisons between gated decoding and unrestricted decoding with twice the canvas. These are not equivalence tests.}
\label{tab:half-canvas}
\begin{tabular}{lccc}
\toprule
Model & Gated at $L$ & Unrestricted at $2L$ & Difference \\
\midrule
LLaDA-8B & $L{=}256$: 0.296 & $L{=}512$: 0.275 & $+2.0\pp$ \\
          & $L{=}512$: 0.356 & $L{=}1024$: 0.348 & $+0.8\pp$ \\
LLaDA-1.5 & $L{=}256$: 0.336 & $L{=}512$: 0.267 & $+6.9\pp$ \\
           & $L{=}512$: 0.336 & $L{=}1024$: 0.336 & $0.0\pp$ \\
Dream-7B & $L{=}256$: 0.417 & $L{=}512$: 0.405 & $+1.2\pp$ \\
          & $L{=}512$: 0.429 & $L{=}1024$: 0.372 & $+5.7\pp$ \\
\bottomrule
\end{tabular}
\end{table}

\section{Controlled answer delay}
\label{app:a2}

\subsection{Four-option construction and scaffold}

Each GSM8K question is converted into a four-option problem. Three distinct distractors are drawn from training-set answers with the same order of magnitude as the correct value. The correct value and distractors are assigned to A, B, C, and D with a fixed per-question permutation. The labels are one tokenizer token. The empirical majority-label baseline is $26.7\%$. Because the distractors come from other questions, the task controls answer position and length but does not reproduce problem-specific arithmetic mistakes.

The scaffold contains 280 masked reasoning positions and one masked answer position. Delimiters and the termination tail are fixed before generation, leaving 281 generated positions. The release step is 140. The evaluation uses the same fixed sample of 640 questions for each model. A train-set interface check gives $99.4\%$ structural compliance. All reported evaluation conditions produce valid A/B/C/D outputs.

\subsection{Matched reasoning delay}

The free evaluation records answer and reasoning commitment times. For each question, the comparison condition selects a reasoning token that commits before step 140 and minimizes the absolute timing difference from the answer. Selection uses commitment times only. All three models have valid matches for all 640 questions; the median and 90th-percentile timing error are one step.

\begin{table}[H]
\centering
\caption{Controlled answer delay ($n{=}640$ for each model).}
\label{tab:a2}
\resizebox{\textwidth}{!}{%
\begin{tabular}{lcccc}
\toprule
Model & Free & Matched reasoning delay & Answer delay & Answer minus reasoning (90\% CI) \\
\midrule
LLaDA-8B & 0.867 & 0.875 & 0.900 & $+2.5\pp$ $[+1.1,+4.1]$ \\
LLaDA-1.5 & 0.875 & 0.884 & 0.906 & $+2.2\pp$ $[+0.6,+3.9]$ \\
Dream-7B & 0.553 & 0.558 & 0.697 & $+13.9\pp$ $[+11.1,+16.7]$ \\
\bottomrule
\end{tabular}%
}
\end{table}

For LLaDA-8B, the 95\% CI is $[+0.8,+4.4]$ and exact McNemar $p{=}0.009$ ($25/9$ discordant pairs). The subset whose free answer commits before release ($98.0\%$) gives $+2.6\pp$ $[+1.1,+4.1]$. LLaDA-1.5 has exact McNemar $p{=}0.044$ ($28/14$ discordant pairs). Dream-7B has exact McNemar $p{=}5.25\times10^{-15}$ ($113/24$ discordant pairs).

\begin{table}[H]
\centering
\caption{Delay-strength check. The matched reasoning condition displaces more top-$k$ selections on every model.}
\label{tab:a2-audit}
\begin{tabular}{lccc}
\toprule
Model & Quantity & Answer delay & Matched reasoning delay \\
\midrule
\multirow{2}{*}{LLaDA-8B} & Median blocked duration & 140 & 140 \\
                           & Median top-$k$ displacement & 81.5 & 115.0 \\
\midrule
\multirow{2}{*}{LLaDA-1.5} & Median blocked duration & 140 & 140 \\
                            & Median top-$k$ displacement & 105 & 123 \\
\midrule
\multirow{2}{*}{Dream-7B} & Median blocked duration & 140 & 140 \\
                           & Median top-$k$ displacement & 77 & 121 \\
\bottomrule
\end{tabular}
\end{table}

\begin{table}[H]
\centering
\caption{How often each delay changes, corrects, or harms the selected option.}
\label{tab:a2-secondary}
\resizebox{\textwidth}{!}{%
\begin{tabular}{llccc}
\toprule
Condition & Outcome & LLaDA-8B & LLaDA-1.5 & Dream-7B \\
\midrule
\multirow{3}{*}{Answer delay}
 & Option changed & 7.2\% & 8.0\% & 24.7\% \\
 & Incorrect $\rightarrow$ correct & 4.5\% & 4.8\% & 18.0\% \\
 & Correct $\rightarrow$ incorrect & 1.2\% & 1.7\% & 3.6\% \\
\midrule
\multirow{3}{*}{Matched reasoning delay}
 & Option changed & 1.9\% & 0.9\% & 5.9\% \\
 & Incorrect $\rightarrow$ correct & 1.2\% & 0.9\% & 2.3\% \\
 & Correct $\rightarrow$ incorrect & 0.5\% & 0.0\% & 1.9\% \\
\bottomrule
\end{tabular}%
}
\end{table}

The total improvement from free to answer delay recovers $24.8\%$, $24.8\%$, and $32.2\%$ of the remaining accuracy headroom for LLaDA-8B, LLaDA-1.5, and Dream-7B. These ratios summarize three models and are not used for inference.

\subsection{Interface dependence}

The scaffold makes the answer position explicit before generation. In this task, the normalized answer commitment step is $0.02$ for LLaDA-8B and $0.007$ for Dream-7B. Their corresponding free-form MATH values are $0.23$ and $0.42$. The task and output format differ, so this is not a matched scaffold intervention. It shows that the controlled task produces substantially earlier answer commitment and limits generalization to ordinary generation. All three A2 conditions share the same scaffold, leaving the answer-versus-reasoning comparison internally matched.

Dream-7B's free accuracy on the controlled task is $0.553$, despite no non-A/B/C/D responses, while its reported MATH accuracy is higher than the LLaDA-8B value. The cause of this cross-task difference is unresolved. LLaDA-1.5 and Dream-7B both commit the scaffolded answer at a median of roughly two denoising steps, yet their raw answer-delay gains differ sharply; timing alone does not explain the cross-model effect size.

\section{Difficulty and per-question timing}
\label{app:difficulty}

On GSM8K, harder problems commit answers slightly later ($\rho=0.144$, $p=0.026$); the association vanishes after controlling for reasoning-region timing. On MATH-500, the raw association is not significant. Difficulty changes the frequency of answer-first responses more clearly than their median timing: the rate rises from $36.4\%$ to $92.2\%$ across MATH levels and from $4.5\%$ to $41.1\%$ across GSM8K gold-step groups. Per-question $t^{*}$ is not used as a universal error score.

\section{Code tasks and return-token timing}
\label{app:tasks}

At the main EOS/EOT setting, frontier gating changes HumanEval by $-1.2\pp$ and MBPP by $-0.4\pp$; neither difference is significant. A HumanEval diagnostic collected under a different EOS setting commits the \texttt{return} token at $t^{*}{=}0.957$ against a function-body median of $\tau{=}0.723$, with $10.3\%$ return-first outputs. We do not compare these timing values directly with the main math conditions because the generation settings differ.

\begin{figure}[H]
\centering
\includegraphics[width=0.90\textwidth]{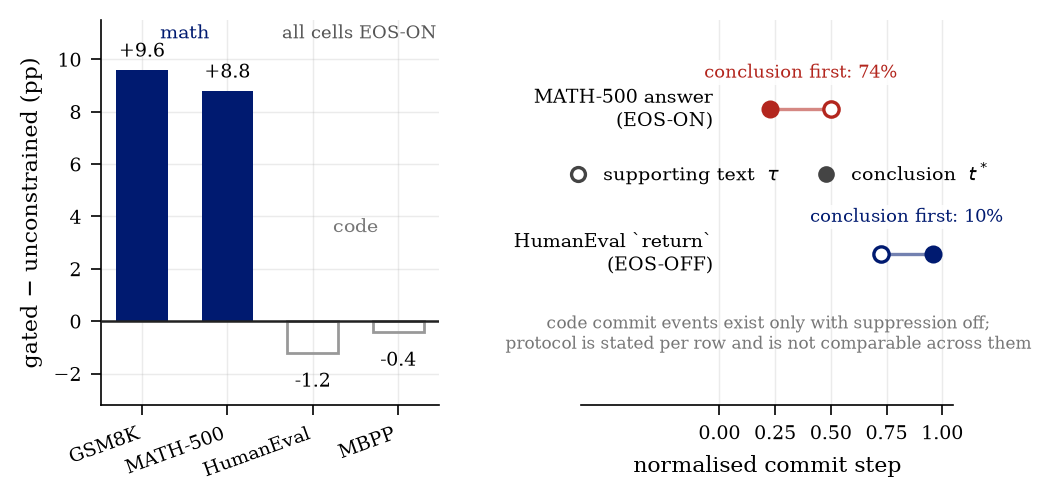}
\caption{\textbf{Accuracy boundary and supplemental timing diagnostic.} Math and code accuracy use the main EOS/EOT setting. The timing rows use different termination settings and are shown only as within-task diagnostics.}
\label{fig:boundary-diagnostic}
\end{figure}

\section{EOS/EOT diagnostic}
\label{app:suppression}

All main results retain the released EOS/EOT confidence handling. The conditions below remove it to inspect termination behavior.

\begin{figure}[H]
\centering
\includegraphics[width=0.86\textwidth]{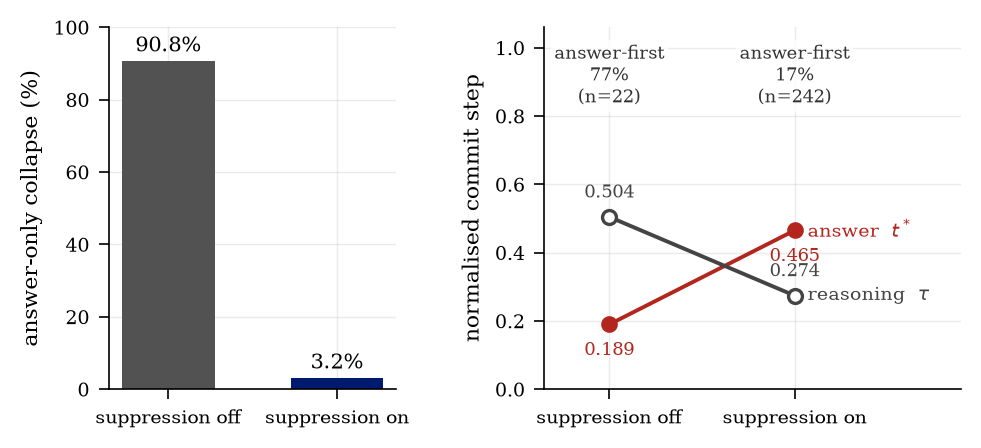}
\caption{\textbf{EOS/EOT suppression on GSM8K.} Disabling suppression produces answer-only collapse and changes commitment order. Timing in the suppression-off condition uses 22 non-collapsed responses.}
\label{fig:suppression}
\end{figure}

\begin{table}[H]
\centering
\caption{EOS/EOT suppression on GSM8K with other settings fixed.}
\label{tab:suppression}
\begin{tabular}{lccccc}
\toprule
Setting & Accuracy & Collapse & $t^{*}$ & $\tau$ & Answer-first \\
\midrule
Suppression off & 0.528 & 90.8\% & 0.189 & 0.504 & 77\% \\
Suppression on  & 0.756 & 3.2\% & 0.465 & 0.274 & 17\% \\
\bottomrule
\end{tabular}
\end{table}

Removing suppression reallocates early commitments to content tokens in addition to changing response length. A supplemental suppression-off prompt--policy interaction is $+34.8\pp$ with a 95\% CI of $[26.8,42.8]$; it is not used in the main argument.

\begin{figure}[H]
\centering
\includegraphics[width=0.70\textwidth]{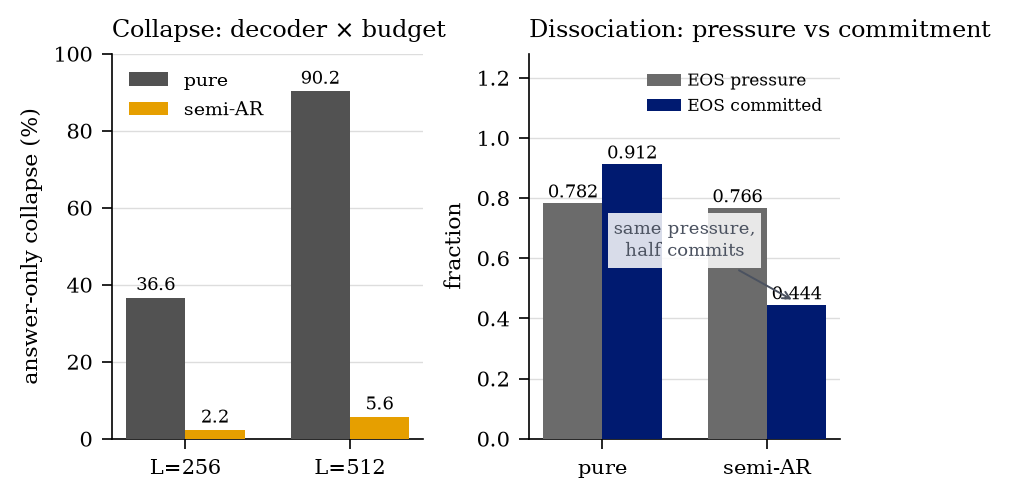}
\caption{\textbf{Termination without EOS/EOT suppression.} Answer-only collapse grows with canvas length under unrestricted decoding and remains lower under semi-autoregressive decoding.}
\label{fig:collapse}
\end{figure}

\section{Implementation checks}
\label{app:repro}

Exact model versions and generation parameters were recorded for every evaluation. Each paired comparison uses one hardware configuration. Instrumentation was checked against the unmodified sampler, and a 40-question regression test compares output tokens, commitment steps, eligible-set sizes, and per-step commitment counts after changes to the intervention code. Reported tables are regenerated from saved outputs by analysis scripts.

At $L{=}512$, frontier gating and semi-autoregressive decoding remain within 1.5\% of unrestricted per-step wall-clock time in this implementation. MATH scoring applies normalized string matching and symbolic verification identically across policies.

\section{Additional diagnostics}
\label{app:generality}

Older Dream-7B evaluations without EOS/EOT suppression give 0.376 accuracy for unrestricted decoding, 0.768 for frontier gating with $w{=}32$, and 0.780 for semi-autoregressive decoding. The large unrestricted termination failure makes these values different from the matched EOS-on MATH comparison in Section~\ref{sec:canvas}.

Adaptive frontier widths preserve full-refinement accuracy but do not repair degradation at small denoising budgets. These evaluations concern termination and commitment density and are not used in the main conclusions.

\end{document}